\documentclass[10pt,twocolumn,letterpaper]{article}

\usepackage[pagenumbers]{cvpr} % To force page numbers, e.g. for an arXiv version
\usepackage{epsfig}
\usepackage{graphicx}
\usepackage{amsmath}
\usepackage{amssymb}
\usepackage{multicol}
\usepackage{multirow}
\usepackage{booktabs}
\usepackage{bbding}
\usepackage{makecell}
\usepackage{subcaption}
\usepackage{caption}
\usepackage[accsupp]{axessibility}
\usepackage{amsmath,amsfonts,bm}

\def\eqref#1{equation~\ref{#1}}
\def\1{\bm{1}}

\DeclareMathAlphabet{\mathsfit}{\encodingdefault}{\sfdefault}{m}{sl}
\SetMathAlphabet{\mathsfit}{bold}{\encodingdefault}{\sfdefault}{bx}{n}

\newcommand{\R}{\mathbb{R}}

\definecolor{cvprblue}{rgb}{0.21,0.49,0.74}
\usepackage[pagebackref,breaklinks,colorlinks,allcolors=cvprblue]{hyperref}

\title{Spatial-Temporal State Propagation Autoregressive Model for 4D Object Generation}

\author{Liying Yang$^{1}$$^{*}$, Jialun Liu$^{2\dagger}$, Jiakui Hu$^{3}$, Chenhao Guan$^{1}$, Haibin Huang$^{2}$, Fangqiu Yi$^{2}$, \\ Chi Zhang$^{2}$, Yanyan Liang$^{1\dagger}$\\
$^{1}$ Macau University of Science and Technology \qquad $^{2}$ TeleAI \qquad $^{3}$ Peking University \\
$^{\dagger}$ Corresponding Authors
}

\begin{document}
\maketitle

\renewcommand{\thefootnote}{\fnsymbol{footnote}}
\footnotetext[1]{Research done when Liying was an intern at TeleAI}

\begin{abstract}
Generating high-quality 4D objects with spatial-temporal consistency is still formidable. Existing diffusion-based methods often struggle with spatial-temporal inconsistency, as they fail to leverage outputs from all previous timesteps to guide the generation at the current timestep. Therefore, we propose a \textbf{S}patial-\textbf{T}emporal State Propagation \textbf{A}uto\textbf{R}egressive Model (4DSTAR), which generates 4D objects maintaining temporal-spatial consistency. 4DSTAR formulates the generation problem as the prediction of tokens that represent the 4D object. It consists of two key components: (1) The dynamic spatial-temporal state propagation autoregressive model (STAR) is proposed, which achieves spatial-temporal consistent generation. Unlike standard autoregressive models, STAR divides prediction tokens into groups based on timesteps. It models long-term dependencies by propagating spatial-temporal states from previous groups and utilizes these dependencies to guide generation at the next timestep. To this end, a spatial-temporal container is proposed, which dynamically updating the effective spatial-temporal state features from all historical groups, then updated features serve as conditional features to guide the prediction of the next token group. (2) The 4D VQ-VAE is proposed, which implicitly encodes the 4D structure into discrete space and decodes the discrete tokens predicted by STAR into temporally coherent 4D Gaussians. Experiments demonstrate that 4DSTAR generates spatial-temporal consistent 4D objects, and achieves performance competitive with diffusion models.
\end{abstract}    
\section{Introduction}
\label{sec:intro}

\begin{figure}[t]%
\centering
\includegraphics[width=0.48\textwidth]{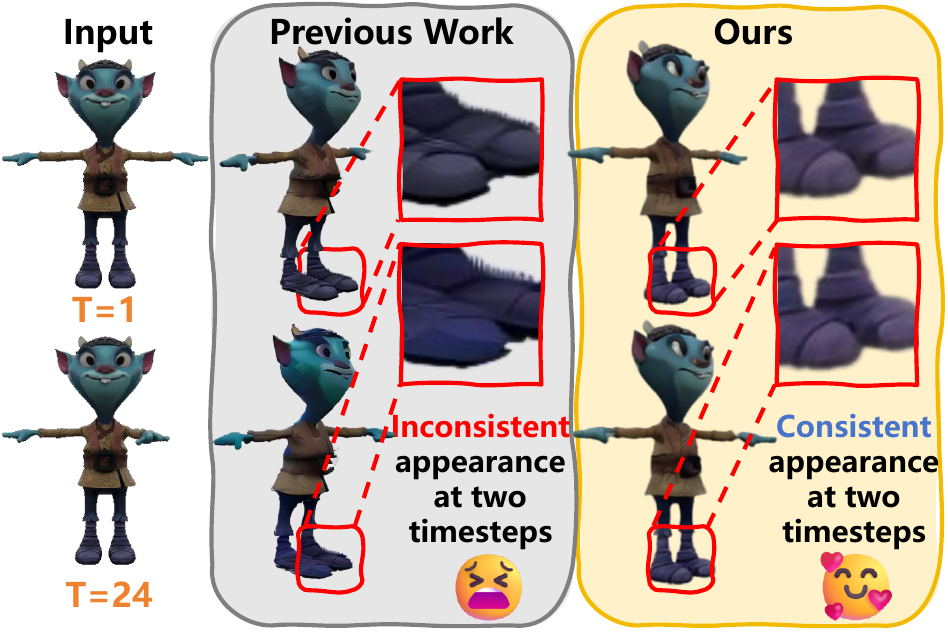}
\caption{Diffusion-based methods, such as previous work~\cite{yao2025sv4d}, fail to leverage outputs from all previous timesteps to guide the generation at the current timestep, which generates results with inconsistent appearance at some timesteps. Our 4DSTAR alleviates this issue by leveraging historical outputs to guide the generation at the current timestep.}\label{teaser}
\end{figure}

Generating 4D object has made tremendous progress. The recent 4D object generation approaches can be called into two main categories. Optimization methods~\cite{singer2023text,4dfy,makeavideo3d,consistent4d,stag4d,yin20234dgen,wu2024sc4d,Yang_2025_ICCV} generate 4D objects by score distilling~\cite{dreamfusion} prior knowledge from pre-trained diffusion models. But the score distillation is sensitive to prompts~\cite{ren2024l4gm}, which limit the application of optimization methods. 

Feed-forward methods~\cite{xie2024sv4d,zhang20244diffusion,ren2024l4gm,diffusion4d,yao2025sv4d,zhang2025gaussian} directly train a diffusion model on 4D datasets. However, they often produce 4D objects with spatial-temporal inconsistency, as they struggle to utilize the outputs from all previous timesteps to guide the generation at the current timestep. In an extreme case (Figure~\ref{teaser}), when generating results over a long time span, previous work only relies on the input video and limited view information. This information fails to adequately assist the model in inferring results that are temporally coherent between timestep 1 and timestep 24.

To address these limitations, we present a novel feed-forward \textbf{S}patial-\textbf{T}emporal State Propagation \textbf{A}uto\textbf{R}egressive Model, named 4DSTAR, which generates temporal-spatial consistency 4D objects. 4DSTAR formulates the 4D generation as the token prediction. It comprises a dynamic spatial-temporal state propagation auto-regressive model (STAR) and a 4D VQ-VAE.

Specifically, we propose dynamic spatial-temporal state propagation autoregressive model (STAR), which models long-term dependencies across generations from previous timesteps via spatial-temporal state propagation, and leverages dependencies to guide next timestep generation. Firstly, it is necessary for the model to accurately observe all changes during the motion of a 4D object, because changes in some areas at a future timestep can refer to corresponding areas at previous timesteps. An intuitive analogy is to pause the object at any timestep and observe it from all angles. Inspired by this concept, STAR divides prediction tokens into multiple groups based on timesteps. 

Then, to model long-term dependencies, we propose a Spatial-Temporal Container (S-T Container) integrated into STAR. It is designed to adaptively update effective spatial-temporal state features from all historical prediction groups. In particular, we assume some token features across different historical groups share similarities in texture and geometry. After merging these similar features, the remaining features constitute the effective spatial-temporal state, which provides the reference context for predicting the next token group. Based on this, the S-T container retrieves tokens from all historical groups and dynamically updates the set of merged features. These updated features then serve as conditional features guiding the prediction of the next token group and are sequentially updated and propagated throughout the autoregressive prediction process of STAR. \textbf{In other words, under $T$ frames, when predicting results at any given $t$ ($t\leq T$), we leverage existing $1$ to $t-1$ historical groups' spatial-temporal state features to predict results at timestep $t$, which enables long-term dependencies across $1$ to $t$ results prediction.}

To decode discrete tokens generated by STAR, we propose a 4D VQ-VAE that implicitly encodes the 4D structure into discrete space and decodes the token sequences into dynamic 3D Gaussians. To preserve temporal stability, our 4D VQ-VAE avoids compression along the temporal axis. The Spatial-Temporal Decoder within 4D VQ-VAE consists of a Static GS Generation and Spatial-Temporal Offset Predictor (STOP). Static GS generation decodes discrete tokens to static Gaussian features. At the same time, STOP jointly leverages cross-frame temporal information from token sequences and static Gaussian features to decode per-timestep Gaussian offsets, thereby establishing explicit point-level correspondence across frames. This enables spatiotemporal fusion to correct static Gaussians into a canonical 4D space at each timestep. Finally, 4D VQ-VAE produces spatial-temporal consistent dynamic 3D Gaussians.

The contributions can be summarized as follows:
\begin{itemize}
\item{To the best of our knowledge, we are the first to propose an autoregressive model for 4D object generation.}
\item{To enforce spatial-temporal consistent generation, we propose a Dynamic Spatial-Temporal State Propagation Autoregressive Model (STAR). It models long-term dependencies across previous predictions via spatial-temporal state propagation to guide the generation at the current timestep.}
\item{To decode the tokens predicted by STAR, we propose a 4D VQ-VAE, which implicitly encodes the 4D structure into discrete space and decodes discrete tokens into temporally coherent dynamic 3D Gaussians.}
\item{Experimental results demonstrate that our method can generate spatial-temporal consistent 4D objects and achieves performance competitive with diffusion models.}
\end{itemize}

\section{Related Works}
\label{sec:related}

\subsection{3D Generation}
Generating high-quality 3D representations, such as meshes~\cite{liu2023one2345,wu2024unique3d,xu2024instantmesh,long2024wonder3d}, NeRFs~\cite{mildenhall2021nerf,gu2023nerfdiff,melas2023realfusion}, and 3D Gaussian splats~\cite{kerbl20233d,szymanowicz2024splatter,zou2024triplane} has become a rapidly evolving research area. Early methods~\cite{shen2023anything,tang2023make,xu2023neurallift,qian2023magic123} distill 2D diffusion priors~\cite{rombach2022high,podell2023sdxl} into 3D via Score Distillation Sampling (SDS)~\cite{poole2022dreamfusion}, enabling text- or image-to-3D synthesis but suffering from inefficiency and view inconsistency caused by per-instance optimization and single-view ambiguity~\cite{long2024wonder3d}. To improve scalability, later works~\cite{long2024wonder3d,liu2023zero,shi2023zero123++,liu2023syncdreamer} decouple the process into multi-view generation followed by 3D reconstruction, often fine-tuning pretrained 2D diffusion models on large 3D datasets~\cite{deitke2023objaverse,deitke2024objaverse}. Although these approaches produce visually appealing results, the reconstructed meshes are often unsuitable for downstream tasks such as animation. More recent methods~\cite{zhang20233dshape2vecset,zhang2024clay,zhao2024michelangelo,hunyuan3d2025} abandon 2D priors and train dedicated 3D generative models from scratch, achieving more accurate and detailed geometry. 

\subsection{4D Generation}
Extending 3D generation into the spatiotemporal domain, 4D generation aims to synthesize dynamic 3D content from text, images, videos, or static assets, requiring consistent geometry and temporally coherent motion. MAV3D~\cite{makeavideo3d} applies score distillation sampling (SDS) from video diffusion models to optimize dynamic NeRFs, while 4D-FY~\cite{4dfy} combines image-, video-, and 3D-aware supervision for better structural fidelity. Consistent4D~\cite{consistent4d}, STAG4D~\cite{stag4d}, SC4D~\cite{wu2024sc4d}, DreamMesh4D~\cite{li2024dreammesh4d} and DS4D~\cite{Yang_2025_ICCV} enhance temporal consistency via frame interpolation and multi-view fusion but remain limited by slow optimization and color oversaturation from SDS~\cite{dreamfusion}. Recent models such as Diffusion4D~\cite{diffusion4d}, 4Diffusion~\cite{zhang20244diffusion}, L4GM~\cite{ren2024l4gm}, GVFDiffusion~\cite{zhang2025gaussian}, and SV4D~\cite{sv4d,yao2025sv4d} employ 4D-aware diffusion to generate orbital views of dynamic assets; however, they struggle with spatial-temporal inconsistency. In contrast, our method solves this by utilizing outputs from all previous timesteps to guide the generation at current timestep.

\subsection{Autoregressive Visual Generation} 
Autoregressive (AR) modeling, pioneered by PixelCNN~\cite{van2016conditional}, formulates image generation as a sequential prediction problem over pixels. Subsequent works such as VQVAE~\cite{van2017neural} and VQGAN~\cite{esser2021taming} extend this paradigm by quantizing image patches into discrete tokens and training transformers to learn AR priors, analogous to language modeling~\cite{brown2020language}. To enhance reconstruction fidelity, RQVAE~\cite{lee2022autoregressive} introduces multi-scale quantization, while VAR~\cite{tian2025visual} reformulates AR modeling as next-scale prediction, markedly improving sampling efficiency. Recent efforts further scale AR transformers for text-conditioned visual generation at large scale~\cite{han2024infinity}. In the 3D domain, early studies such as PolyGen~\cite{nash2020polygen} and MeshGPT~\cite{siddiqui2024meshgpt} adopt AR models to directly generate mesh vertices and faces. However, these methods are confined to geometric representations and face challenges in scaling to diverse 3D object datasets~\cite{deitke2023objaverse}. In contrast, we propose an autoregressive model for 4D object generation, which updates effective spatial-temporal state features provided by all historical groups, enabling temporally coherent 4D generation.

\begin{figure*}[!t]%
\centering
\includegraphics[width=1.\textwidth]{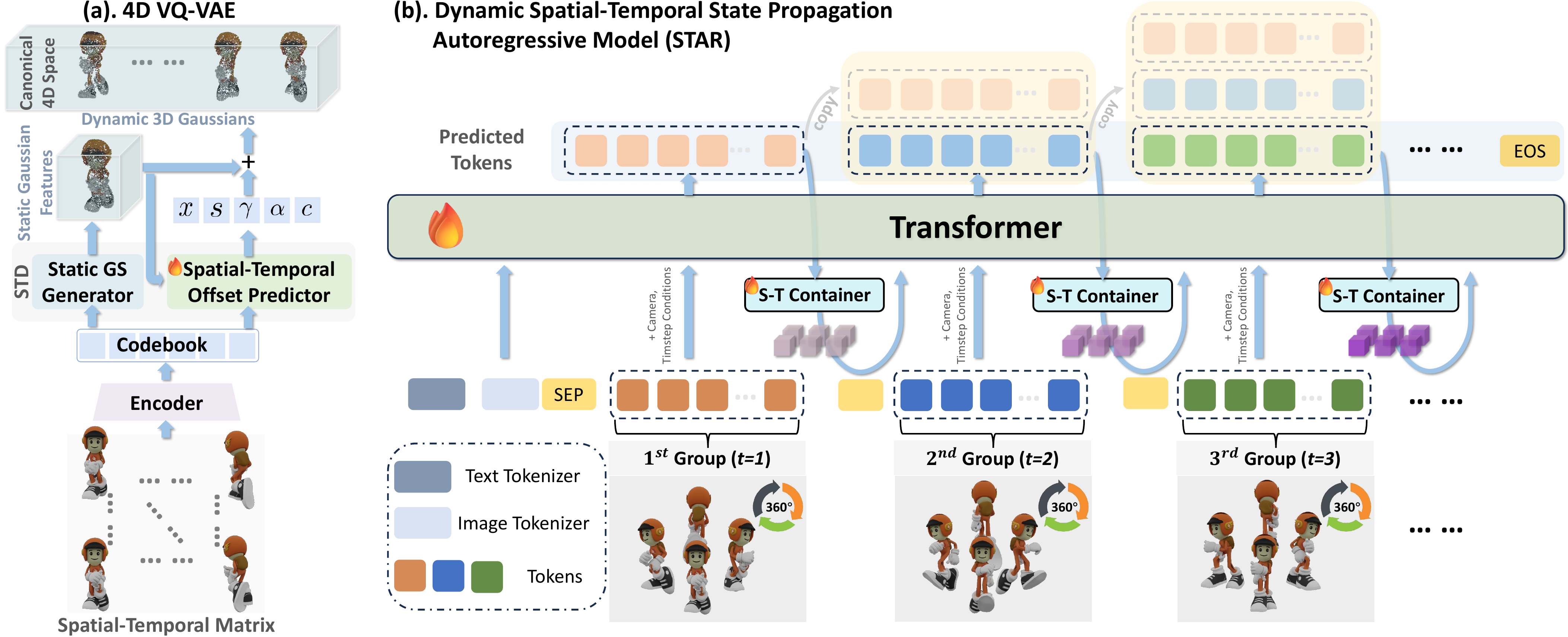}
\caption{\textbf{The overall pipeline of our 4DSTAR.} 4DSTAR consists of two key components: (a) 4D VQ-VAE. Given a 4D object, we first render it as a spatial-temporal matrix. Then the matrix is encoded by Encoder, and is compressed into discrete tokens. Static GS Generation decodes these tokens to static Gaussians. Meanwhile, Spatial-Temporal Offset Predictor (STOP) corrects static Gaussians into a canonical 4D space at each timestep. Finally, the model outputs dynamic 3D Gaussians.  (b) Dynamic Spatial-Temporal State Propagation Autoregressive Model (STAR). The text and video conditions, which are compressed by an image tokenizer are concatenated before the start token as the context. The conditions can either expect the model to generate tokens. The SEP signals the model to begin generating tokens. Then, camera pose and timestep conditions are integrated. When the model starts to predict the next group, the historical groups are integrated into Spatial-Temporal Container (S-T Container). S-T Container updates effective spatial-temporal state features. The features serve as conditional features to guide the prediction of the next token group. Finally, STAR predicts all tokens that represent a 4D object.}\label{pipeline}
\end{figure*}
\section{Methodology}

We aim to generate high-quality 4D objects with spatial-temporal consistency. Figure.~\ref{pipeline} shows the overview of 4DSTAR, including 4D VQ-VAE and dynamic Spatial-Temporal State Propagation auto-Regressive model. 

\subsection{4D Vector Quantized Variational Autoencoder} \label{4dvqvae} 

In our 4D VQ-VAE, each 4D object is regarded as the spatial-temporal matrix, which is arranged by 2D view images $x \in \R^{T \times V \times H \times W \times 3}$, where $T$ denotes the number of frames, $V$ denotes the number of views. To capture both fine-grained details and high-level semantics within each 2D view images, we employ the encoder of UniTok~\cite{ma2025unitok}. The encoder encodes the spatial-temporal matrix to latent vector, and then obtain the corresponding discrete tokens. Specifically, the matrix is quantized into $q\in Q^{T\times V\times n\times\frac{d}{n}}$, where $n$ is the dimension of latent vector, $n$ denotes the number of chunks which used to split latent vector.

The discrete tokens extracted from the matrix require a decoder for reconstruction. Although a straightforward solution is to employ the UniTok decoder, it suffers from a fundamental limitation due to the gap between 2D and 4D representations. Specifically, UniTok decoder is designed for independent 2D image reconstruction and fails to capture the inherent geometric constraints in 4D. Consequently, this leads to spatial-temporal inconsistencies across the reconstructed frames. To address this issue, we propose a spatial-temporal decoder (STD). In contrast to UniTok decoder, STD decodes tokens as dynamic 3D Gaussians.

The STD commits to reconstruct 4D results represented by 4D Gaussian splatting. Specifically, for input discrete tokens $q$, we employ the multi-head attention modules for token factorization projection. It converts the discrete tokens into continuous tokens $S \in \R^{T\times V\times C\times N}$, where $C$ denotes the dimension of tokens, $N$ denotes the number of tokens. Then, through encoding and decoding these continuous tokens, the coarse decoder in STD can directly decode coarse Gaussians $G^t=\{g_1^t,g_2^t,\dots,g_n^t\}$, with the parameters $g_i^t=\{x_i^t,s_i^t,\gamma_i^t,\alpha_i^t,c_i^t\}$ for each timestep $t$. To further correct these static Gaussians into a canonical 4D space at each timestep, we propose Spatial-Temporal Offset Predictor (STOP) inserted into STD, which jointly leverages cross-frame temporal information from token sequences and static GS features.

\textbf{Spatial-Temporal Offset Predictor (STOP).} Specifically, we input static Gaussians and continuous tokens $\textbf{t}$ of each object into the STOP. For the dynamic 3D Gaussians $\textbf{G}=\{G^1,G^2,\dots,G^T\}$ in the whole timesteps $T$, and continuous tokens $\textbf{S}_v=\{S_v^1,S_v^2,\dots,S_v^T\}$ belonging to the whole timesteps $T$ at each view $v$, we calculate the cross-attention among static Gaussian features and continuous tokens along timestep axes. In detail, we set the $\textbf{G}$ as query and $\textbf{S}_v$ as key and value. During the cross-attention calculation, each static Gaussian features can aggregate the global temporal information based on the similar geometry and texture relationships among continuous tokens in temporal axes, thereby leveraging temporal context to obtain the coarse offset features at each timestep. Then, a 3D-Unet module is used to refine the coarse offset features, which enhance the 3D awareness of the features. The refined offset features $f_o^t=\{x_{io}^t,s_{io}^t,\gamma_{io}^t,\alpha_{io}^t,c_{io}^t\}$ can be split as the offset value of each parameter of the dynamic 3D Gaussians. At last, we update the parameters of dynamic 3D Gaussians as $\hat{g_i}^t=\{x_i^t+x_{io}^t,s_i^t+s_{io}^t,\gamma_i^t+\gamma_{io}^t,\alpha_i^t+\alpha_{io}^t,c_i^t+c_{io}^t\}$. 

\textbf{Loss Function.} We employ pixel-level rendering loss $\mathcal{L}_R$ among ground truth and rendering views, and use a discriminator loss $\mathcal{L}_G$ to enhance reconstruction fidelity~\cite{karras2019style}. Furthermore, we introduce the optical flow loss $\mathcal{L}_F$ to guide the motion modeling. For 4D VQ-VAE training, the overall loss function $\mathcal{L}_{\text{VAE}}=\alpha\mathcal{L}_R+\beta\mathcal{L}_G+\gamma\mathcal{L}_F$, where $\alpha, \beta, \gamma$ is weight. More details see supplementary materials.

\subsection{Dynamic Spatial-Temporal State Propagation Auto-Regressive Model (STAR)}

In Sec.~\ref{4dvqvae}, we introduce how our 4D VQ-VAE encodes and decodes the tokens representing 4D objects. In this section, we introduce our STAR how to predict these tokens.

To build long-term dependencies, a simple approach is to use a standard auto-regressive model (e.g., LlamaGen~\cite{sun2024autoregressive}) to predict these tokens. However, this approach faces significant challenges. The primary difficulty is the need to precisely predict a massive sequence of tokens (over 40,000) that encapsulates a 4D object. This challenge is compounded by high token density, which complicates accurate prediction, and the fact that not all historical information is useful for forecasting the next token. To address these limitations, we propose a Dynamic Spatial-Temporal State Propagation Auto-Regressive Model (STAR), which updates effective spatial-temporal state features based on historical groups. In this section, we provide a detailed introduction to STAR.

\subsubsection{Conditions} \label{cond}
Firstly, we introduce the conditions integrated into STAR.

\textbf{Text Condition.} To extract text features, we utilize FLAN-T5 XL~\cite{chung2024scaling} as the text encoder. Then, the text features are projected by a text tokenizer, which contains an MLP, and serve as the prefill token embeddings in STAR.

\textbf{Camera Condition.} To condition the camera pose within a group, following ~\cite{kant2024spad,hu2025auto}, we apply the Pl{\"u}cker Embedding. This embedding encodes the origin and direction of the ray at each spatial location. The ray is then incorporated into STAR as the Shift Positional Encoding (SPE).

\textbf{Timestep Condition.} To condition the temporal information across groups, we design a Timestep encoder that maps each Timestep $t$ to a temporal embedding $\hat{t} \in \R^{D_p}$, with $D_p$ aligning the dimensions of the Pl{\"u}cker Embedding. Similar to to the latter, this temporal embedding is incorporated into STAR as the SPE.

\textbf{Monocular Video Condition.} Given a monocular video of $T$ frames, we represent it as the set $\textbf{F}=\{F_1, F_2, \dots, F_T\}$. The 4D VQ-VAE encoder first converts this input into discrete tokens. These tokens are then projected by an image tokenizer, which contains an MLP, to form the prefill token embeddings for STAR.

\subsubsection{Network Architecture}
Based on the conditions in Sec.~\ref{cond}, STAR is committed to predicting the discrete tokens that represent a 4D object. 

\textbf{Dividing Groups.} First, building upon the concept introduced in Sec.~\ref{sec:intro}, STAR divides the prediction tokens into multiple groups, as illustrated in Figure~\ref{pipeline} (b). Specifically, we consider a prediction token sequence $Q=\{q_1^1,q_1^2,q_1^3,\dots,q_t^v,\dots,q_T^V\}$, where $T$ denotes the number of timesteps and $V$ denotes the number of views. This sequence is partitioned into $T$ groups, denoted as $\textbf{L}=\{L_1, L_2, \dots, L_T\}$. The group $L_t$ at time $t$, contains the partial prediction tokens $L_t=\{q_t^1,q_t^2,\dots,q_t^V\}$. Thus, all prediction tokens are organized into groups, which STAR then predict them sequentially. During training, the groups $\textbf{L}$ are projected by the image tokenizer into token features $\hat{\textbf{L}}=\{\hat{L}_1,\hat{L}_2,\dots,\hat{L}_T\}$. Then the token features and conditions serve as the input to STAR. During inference, STAR autoregressively predicts the token sequences for each group based on the conditioning inputs.

\textbf{Spatial-Temporal Container.} We propose the Spatial-Temporal Container (S-T Container) to build long-term dependencies by dynamically updating spatial-temporal state features from historical groups. S-T Container is designed to merge token features that exhibit similar texture and geometry across historical groups. To identify and extract these potentially similar token features, we employ a k-nearest neighbor based density peaks clustering algorithm (DPC-KNN)~\cite{du2016study}.

Specifically, given token features $\hat{\textbf{L}}_{t\in[1,t-1]}$ from group 1 to group $t-1$ (i.e., time 1 to time $t-1$), we compute the local density $\rho_i$ of each token within groups:

\begin{equation}
  \label{equ:knn}
  \rho_i^m= \mathtt{exp}(-\frac{1}{K}\sum_{y_j^n\in\mathtt{KNN}(y_i^m)}\Vert y_i^m-y_j^n\Vert^2_2),
\end{equation}

where $\mathtt{KNN}(y_i^m)$ denotes the k-nearest neighbors of the $m$-th token feature in $i$-th group. Next, we compute the similar score $\varpi$ as the minimal distance among token features with higher local density:

\begin{equation}
  \label{equ:knn2}
    \varpi_i^m =\begin{cases}
    \mathop{min}\limits_{j,n:\rho_j^n>\rho_i^m}\Vert y_i^m-y_j^n \Vert^2_2, \quad \text{if}\ \exists j,n \ \text{s.t}\ \ \rho_j^n>\rho_i^m.  \\
    \ \ \ \ \mathop{max}\limits_{j,n}\Vert y_i^m-y_j^n \Vert^2_2, \quad \text{otherwise}. \\
    \end{cases}
\end{equation}

where $\rho_i^m$ denotes local density. Then we set cluster centers based on the score $\rho_i^m \times \varpi_i^m$. A higher score means a higher possibility of being a cluster center.

Inspired by feature fusion~\cite{xie2019pix2vox,rao2021dynamicvit,zeng2022not}, we use an MLP to predict the dissimilar score $\sigma$ of each token feature. Then we merge the token features according to $\sigma$, the merged token features $\hat{y}_i=\sum_{j\in C_i}\sigma_jy_j$, where $C_i$ denotes the set of the $i$-th cluster, and $y_j$ means the $j$-th token features in the $i$-th cluster. Next, we refine the merged token features via multi-head attention, obtaining the updated features $\hat{y}_i$.

The updated features $\hat{y}_i$ represents the effective spatial-temporal state features provided by groups 1 to $t-1$. These features then serve as condition features, integrated into STAR via MLPs, for predicting the tokens in group $t$. \textbf{\textit{As the prediction progresses through the groups, the S-T container iteratively updates the refined token features. In other words, it gradually propagates the dynamic spatial-temporal state, thereby building long-term dependencies across all timesteps.}}

\textbf{Transformer Architecture.} We employ Transformer for auto-regressive modeling, which is consisted with standard auto-regressive model. The Transformer is developed based on Llama~\cite{touvron2023llama}.

\textbf{Loss Function.} Following auto-regressive models~\cite{touvron2023llama,sun2024autoregressive}, our STAR generates the conditional probability $p(q_s|q_{<s})$ of token $q_s$ at each position $s$. Then, the cross-entropy (CE) loss is defined as the average of the negative log-likelihoods over all vocabulary positions:

\begin{equation}
    \mathcal{L}_{ar}=-\frac{1}{S}\sum_{s=1}^S\mathtt{log}p(q_s|q_{<t}), \label{ar_func1}
\end{equation}

The Eq.~\ref{ar_func1} indicates that STAR studies the transformation process from previous $s-1$ tokens to $s$-th tokens. Due to we have $T$ groups in STAR, we employ chunked CE loss, which we split the chunk based on the group. Each chunk includes one group, formally,

\begin{equation}
    \mathcal{L}_{AR}=\sum_{g=1}^G\mathtt{Chunk}(L_{ar}^1,L_{ar}^2,\dots,L_{ar}^G).
\end{equation}
 where $G$ means the number of groups.

\section{Experiments} \label{exps}

\begin{figure*}[!t]%
\centering
\includegraphics[width=1.\textwidth]{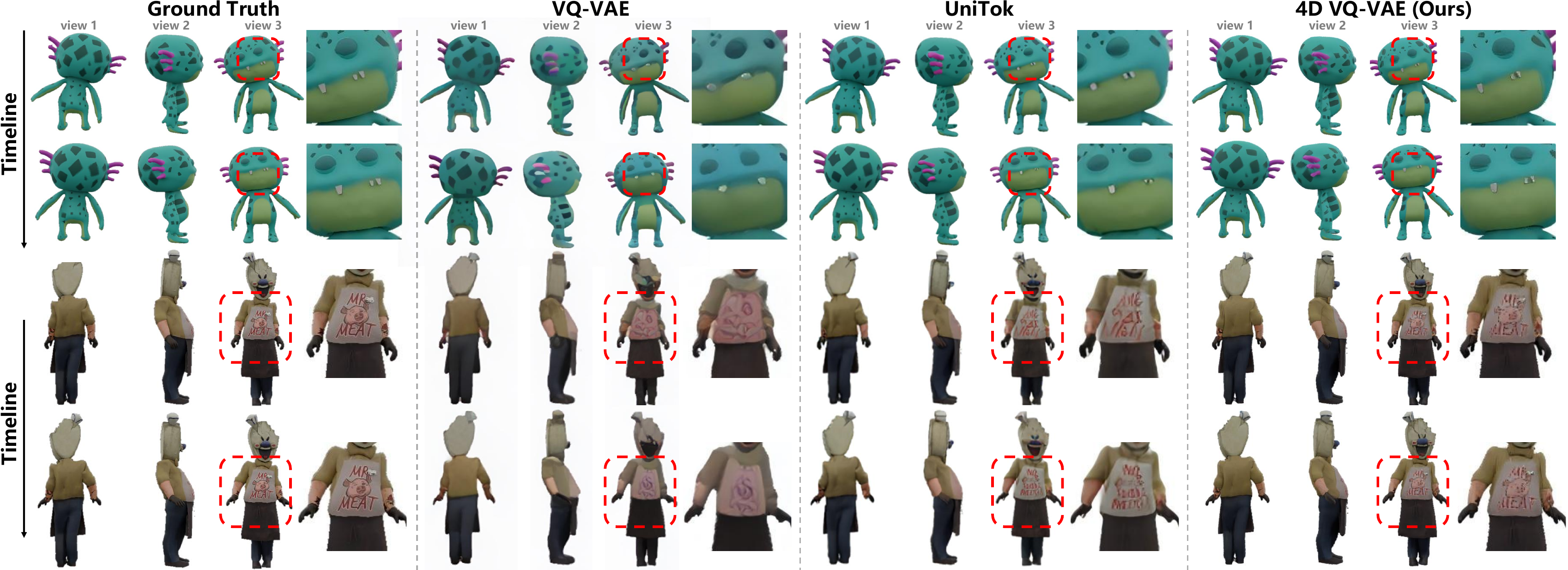}
\caption{\textbf{Qualitative comparison with VQ-VAE~\cite{sun2024autoregressive} and UniTok~\cite{ma2025unitok} on 4D reconstruction.} For VQ-VAE and UniTok, we employ them to reconstruct 2D view images. For our 4D VQ-VAE, we render results under corresponding views. Our 4D VQ-VAE can reconstruct the results with temporal coherence, while VQ-VAE and UniTok cannot reconstruct with temporal coherence.}\label{vae_exp}
\end{figure*}

\begin{figure*}[!t]%
\centering
\includegraphics[width=1.\textwidth]{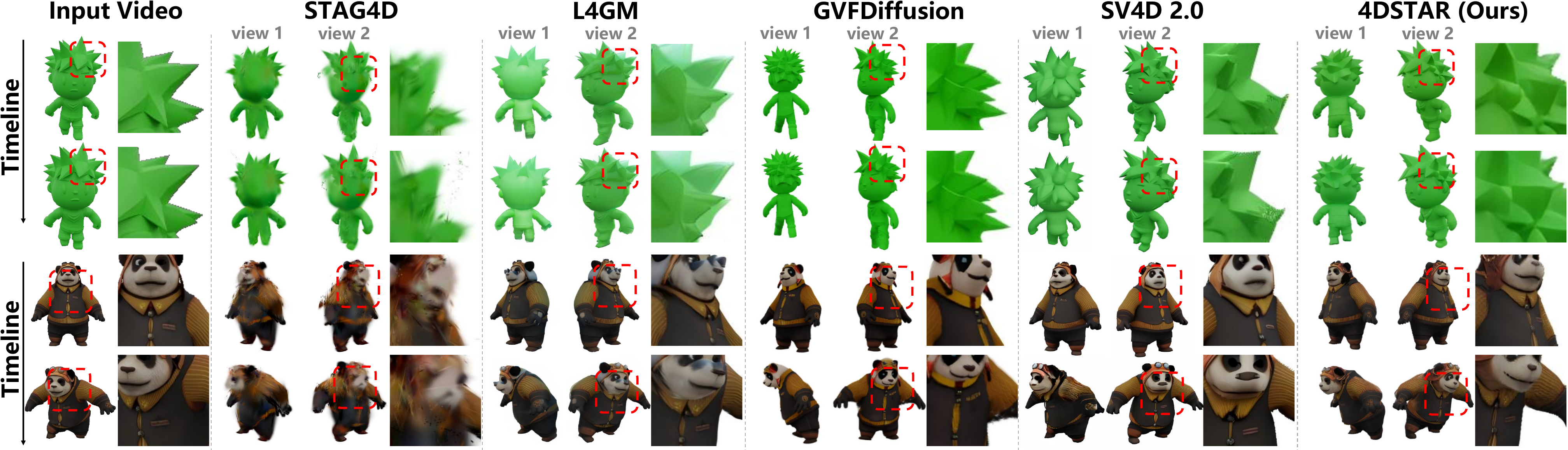}
\caption{\textbf{Qualitative comparison with the state-of-the-art methods~\cite{stag4d,ren2024l4gm,zhang2025gaussian,yao2025sv4d} on video-to-4D generation.} For each method, we render results under two novel views at two timesteps. Our method achieves high-quality generation with spatial-temporal consistency.}\label{ar_exp}
\end{figure*}

\subsection{Implementation Details}
\textbf{Training Dataset and Metrics.} We train 4D VQ-VAE and STAR on the train set of Objaverse~\cite{deitke2023objaverse} and Objaverse-XL~\cite{deitke2024objaverse}, which includes 56K 4D objects. For text prompts, we first utilize those provided by Cap3D~\cite{luo2023scalable}. However, as these prompts describe static appearance, we employ Qwen~\cite{bai2025qwen2} to generate additional prompts that capture object motion. To evaluate the quality of our 4D generation, we adopt metrics from~\cite{consistent4d, stag4d}, including CLIP, LPIPS, FVD, and FID-VID. The same metrics are also used to evaluate the reconstruction quality of 4D VQ-VAE. All evaluations are conducted on renderings at a resolution of $512\times512$.

\textbf{Training.} We train our 4D VQ-VAE on 8 H100 GPUs and STAR on 16 H100 GPUs. More details can be found in the Supplementary Materials.

\subsection{Main Results}

\subsubsection{4D Object Reconstruction}

\textbf{Evaluation Dataset.} To evaluate the performance on 4D object reconstruction, we construct a test set of 100 objects from Objaverse and Objaverse-XL. We render 8 views of each timestep for each object. These views are grouped as a spatial-temporal matrix as the input of each method. Besides, we select 4 views (front, back, left, and right) of each timestep for each object, which are used as ground truth.

\textbf{Quantitative comparisons.} We compare our 4D VQ-VAE with two 2D image VQ-VAE, including VQ-VAE~\cite{sun2024autoregressive} and UniTok~\cite{ma2025unitok}. As shown in Table.~\ref{abl_vae_table}, our 4D VQ-VAE outperforms other VQ-VAE across all quality metrics. It demonstrates that our VQ-VAE has both superior reconstruction fidelity and better temporal coherence. This is attributed to our 4D VQ-VAE leveraging spatial-temporal information within tokens to reconstruct results.

\begin{table}[]
	\centering
	\renewcommand\arraystretch{0.95}
	\scalebox{0.8}{
    \begin{tabular}{c|cccc}
	    \Xhline{1.5px}
		\multicolumn{1}{c|}{\textbf{\makecell{Method}}} &\multicolumn{1}{c}{\textbf{CLIP $\uparrow$}} & \multicolumn{1}{c}{\textbf{LPIPS $\downarrow$}} &
		\multicolumn{1}{c}{\textbf{FVD $\downarrow$}} & \multicolumn{1}{c}{\textbf{FID-VID $\downarrow$}}  \\ \hline
          \textbf{VQ-VAE~\cite{sun2024autoregressive}} & 0.938 & 0.067 & 326.570 & 13.447  \\ 
          \textbf{UniTok~\cite{ma2025unitok}} & \textbf{0.973} & 0.050  & 151.238 & 4.402  \\ \hline % mvar的vqgan
		\textbf{4D VQ-VAE (Ours)} & \textbf{0.973} & \textbf{0.048} & \textbf{133.372} & \textbf{4.175}  \\
        \Xhline{1.5px}
	\end{tabular}}
	\caption{\textbf{Evaluation and comparison of the performance on 4D object reconstruction.} The best score is highlighted in bold. All the experiments of the methods are carried out using the code from their official GitHub repository.}
\label{abl_vae_table}
\end{table}

\begin{table}[]
	\centering
	\renewcommand\arraystretch{0.95}
	\scalebox{0.8}{
    \begin{tabular}{c|cccc}
	    \Xhline{1.5px}
		\multicolumn{1}{c|}{\textbf{\makecell{Method}}} &\multicolumn{1}{c}{\textbf{CLIP $\uparrow$}} &
		\multicolumn{1}{c}{\textbf{LPIPS $\downarrow$}}  & \multicolumn{1}{c}{\textbf{FVD $\downarrow$}} & \multicolumn{1}{c}{\textbf{FID-VID $\downarrow$}} \\ \hline
		  \textbf{STAG4D~\cite{stag4d}}  & 0.910 & 0.141 & 752.215 & 27.280 \\
          \textbf{L4GM~\cite{ren2024l4gm}}   & 0.926 & 0.132  & 515.430 & 20.125 \\
          \textbf{SV4D 2.0~\cite{yao2025sv4d}}  & 0.932 & 0.139 & 497.753  & 19.223 \\
          \textbf{GVFDiffusion~\cite{zhang2025gaussian}}  & 0.931 & 0.138 & 528.336 & 19.064  \\ \hline
		\textbf{4DSTAR (Ours)}  & \textbf{0.952} & \textbf{0.131} & \textbf{464.709} & \textbf{15.312} \\ 
        \Xhline{1.5px}
	\end{tabular}}
	\caption{\textbf{Evaluation and comparison of the performance on video-to-4D object generation.} The best score is highlighted in bold. All the experiments of the methods are carried out using the code from their official GitHub repository.}
\label{v24d_table}
\end{table}

\textbf{Qualitative comparisons.} The qualitative results on test set are presented in Figure.~\ref{vae_exp}. Obviously, the reconstruction quality and temporal coherence of our 4D VQ-VAE both are better than other methods. Specifically, for the top object, VQ-VAE and UniTok cannot reconstruct the texture details of eye at two timesteps, while our 4D VQ-VAE successfully reconstruct the details. Moreover, for the below object, VQ-VAE and UniTok reconstruct significantly inconsistent texture results at different timesteps, especially for the clothing texture. In contrast, our 4D VQ-VAE reconstructs consistent texture results at both timesteps, which once again demonstrates the ability of our 4D VQ-VAE in ensuring temporal coherence in reconstruction.

\subsubsection{Video-to-4D Object Generation}

\textbf{Evaluation Dataset.} We construct a test set of 100 objects by combining 7 objects from Consistent4D~\cite{consistent4d} and 93 additional objects from Objaverse-XL test set. To ensure fair comparisons, our model employs a neutral text prompt, ``Generate object of the following $<\text{imgs}>$", which intentionally omits any specific appearance or motion descriptions. For previous works, we follow their original papers and official implementations, using their semantic text embeddings (if they have) instead of neutral text prompt.

\textbf{Quantitative comparisons.} We compare our model with the SOTA methods on the test set, including STAG4D~\cite{stag4d}, L4GM~\cite{ren2024l4gm}, GVFDiffusion~\cite{zhang2025gaussian}, and SV4D 2.0~\cite{yao2025sv4d}. The quantitative results are shown in Table.~\ref{v24d_table}. Our 4DSTAR consistently outperforms other methods in all metrics. Specifically, our 4DSTAR notably exceeds the other methods in FVD and FID-VID, which indicates the results generated by our 4DSTAR have fewer temporal artifacts and better temporal coherence than others. Furthermore, our 4DSTAR significantly outperforms other methods in CLIP. It demonstrates our methods are superior in terms of quality, fidelity, and spatial-temporal consistency of generation results. In summary, these improvements are attributed to our method, which models long-term dependencies across outputs from previous timesteps to enhance spatial-temporal consistent generation.

\textbf{Qualitative comparisons.} The qualitative results are presented in Figure.~\ref{ar_exp}. Visibly, the generation fidelity and spatial-temporal consistency are both better than other methods. Specifically, when generating the results in two timesteps, the results generated by STAG4D, L4GM, and SV4D 2.0 have different degrees of blurriness and inconsistent appearance in the details, especially in areas with complex topology (e.g., boy's hair). Moreover, when generating the results that contain large motion at two timesteps, the results generated by L4GM, GVFDiffusion, and SV4D 2.0 show inconsistent appearance, temporal incoherence, and some noisy points within motion parts. For example, for the arms of the bear, these methods generate clear textures in previous timesteps, while they generate low-quality textures in the future timestep. These methods fail to leverage outputs from previous timesteps to guide generation at the current timestep. In contrast, our 4DSTAR alleviates these issues by leveraging long-term dependencies to guide the generation, achieving high-quality generation.

\subsection{Ablation Study}
\begin{figure}[t]%
\centering
\includegraphics[width=0.44\textwidth]{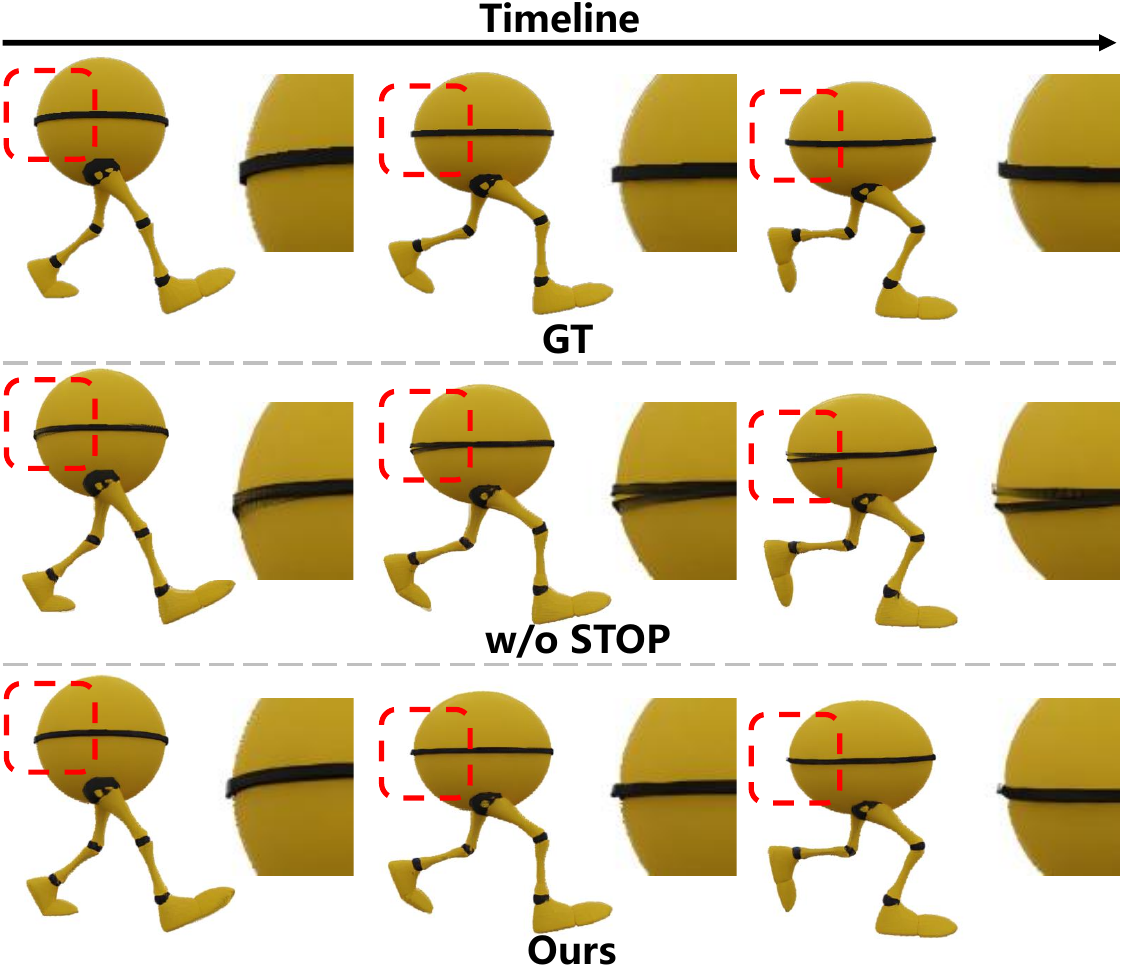}
\caption{\textbf{Ablation of 4D VQ-VAE.} Our model which uses STOP, accurately recovers texture details at different timesteps.}\label{vae_abl}
\end{figure}

\textbf{Ablation of 4D VQ-VAE.} To validate the effect of STOP in 4D VQ-VAE, we conduct an experiment about whether to use STOP in model. The results are shown in Table.~\ref{abl_vae_table}. Our method significantly outperforms the model without using STOP in all metrics, especially FVD and FID-VID. Besides, Figure.~\ref{vae_abl} shows the reconstruction results by different models at three timesteps. The model without STOP cannot reconstruct consistent texture details across timesteps, while our model, which uses STOP, accurately recovers texture details at different timesteps. It indicates the STOP leverage of spatial-temporal information among tokens to correct static Gaussians into a canonical 4D space for better temporally coherent reconstruction.

\begin{table}[]
	\centering
	\renewcommand\arraystretch{0.95}
	\scalebox{0.8}{
    \begin{tabular}{c|cccc}
	    \Xhline{1.5px}
		\multicolumn{1}{c|}{\textbf{\makecell{Method}}} &\multicolumn{1}{c}{\textbf{CLIP $\uparrow$}} & \multicolumn{1}{c}{\textbf{LPIPS $\downarrow$}} &
		\multicolumn{1}{c}{\textbf{FVD $\downarrow$}} & \multicolumn{1}{c}{\textbf{FID-VID $\downarrow$}}  \\  \hline
          \textbf{w/o STOP} & 0.968 & 0.057 & 174.757 & 5.575 \\ \hline % 这其实是lgm的
		\textbf{4D VQ-VAE (Ours)} & \textbf{0.973} & \textbf{0.048} & \textbf{133.372} & \textbf{4.175} \\
        \Xhline{1.5px}
	\end{tabular}}
	\caption{\textbf{Ablation Experiments of 4D VQ-VAE on the test set.} Each setup is based on a modification of the immediately preceding setups. The best score is highlighted in bold.}
\label{abl_vae_table}
\end{table}

\textbf{Ablation of STAR.} To validate the effect of S-T Container in STAR, we conduct several experiments, as shown in Table.~\ref{abl_ar_table}. Among them, model A based on L4GM with explicit state propagation. Baseline is the naive auto-regressive model. Model B, based on baseline, performs average pooling of historical groups along temporal to merge historical groups, while model C uses MLP to merge them. Our model with S-T Container significantly outperforms other models in all metrics. Our model leverages clustering of S-T Container to preserve diverse spatial-temporal information within historical token features. This process provides a crucial inductive bias for modeling long-term dependencies. However, model B and C struggle to accurately extract such information. Although model A explicitly propagates previous results to assist prediction at the next timestep, it only uses the results of one historical frame to guide the generation at the next timestep. This not only makes the model easily forget valid information from all historical results but also fails to filter useful information from historical results, ultimately leading to the generation of spatial-temporal inconsistencies that affect time coherence. In contrast, our model can consistently produce temporally consistent results. By merging regionally similar textures and topological structures via S-T container and propagating remaining features to constitute the effective spatial-temporal state across timesteps, our model gradually establishes long-term dependencies, thereby improving temporal consistency in generation. For model details and more visualizations, please see supplementary materials.

\begin{table}[]
	\centering
	\renewcommand\arraystretch{1.}
	\scalebox{0.63}{
    \begin{tabular}{c|cccc}
	    \Xhline{1.5px}
		\multicolumn{1}{c|}{\textbf{\makecell{Method}}} &\multicolumn{1}{c}{\textbf{CLIP $\uparrow$}} &
		\multicolumn{1}{c}{\textbf{LPIPS $\downarrow$}}  & \multicolumn{1}{c}{\textbf{FVD $\downarrow$}} & \multicolumn{1}{c}{\textbf{FID-VID $\downarrow$}} \\ \hline 
        \textbf{A: L4GM w/ Explicit State Propagation} & 0.930 & 0.132  & 508.192 & 19.714  \\ \hline 
        \textbf{Baseline} & 0.939 & 0.148  & 883.946 & 37.066  \\  % llamagen
        \textbf{B: w/ Temporal Average Pooling} & 0.942 & 0.146 & 646.558 & 27.016  \\ % 
        \textbf{C: w/ Learnable Token Merging} & 0.945 & 0.141 & 579.288 & 21.449  \\ \hline % 
          \textbf{Ours: w/ S-T Container} & \textbf{0.952} & \textbf{0.131} & \textbf{464.709} & \textbf{15.312} \\  % 最终模型
        \Xhline{1.5px}
	\end{tabular}}
	\caption{\textbf{Ablation Experiments of STAR on the test set.} Each setup is based on a modification of the immediately preceding setups. The best score is highlighted in bold.}
\label{abl_ar_table}
\end{table}

\begin{figure}[t]%
\centering
\includegraphics[width=0.45\textwidth]{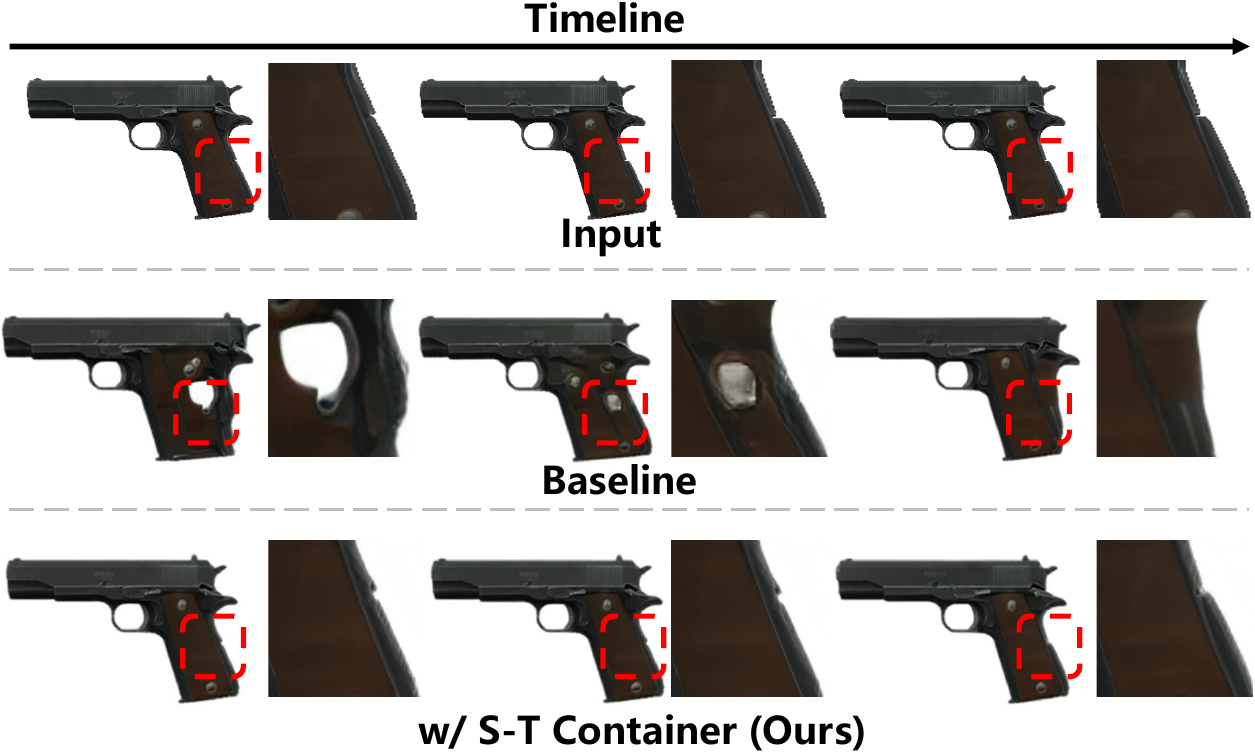}
\caption{\textbf{Ablation of STAR.} Even slight motion within input, ours produces more temporally consistent results than Baseline. More visualizations see supplementary materials.}\label{ar_abl}
\end{figure}

\begin{figure}[t]%
\centering
\includegraphics[width=0.5\textwidth]{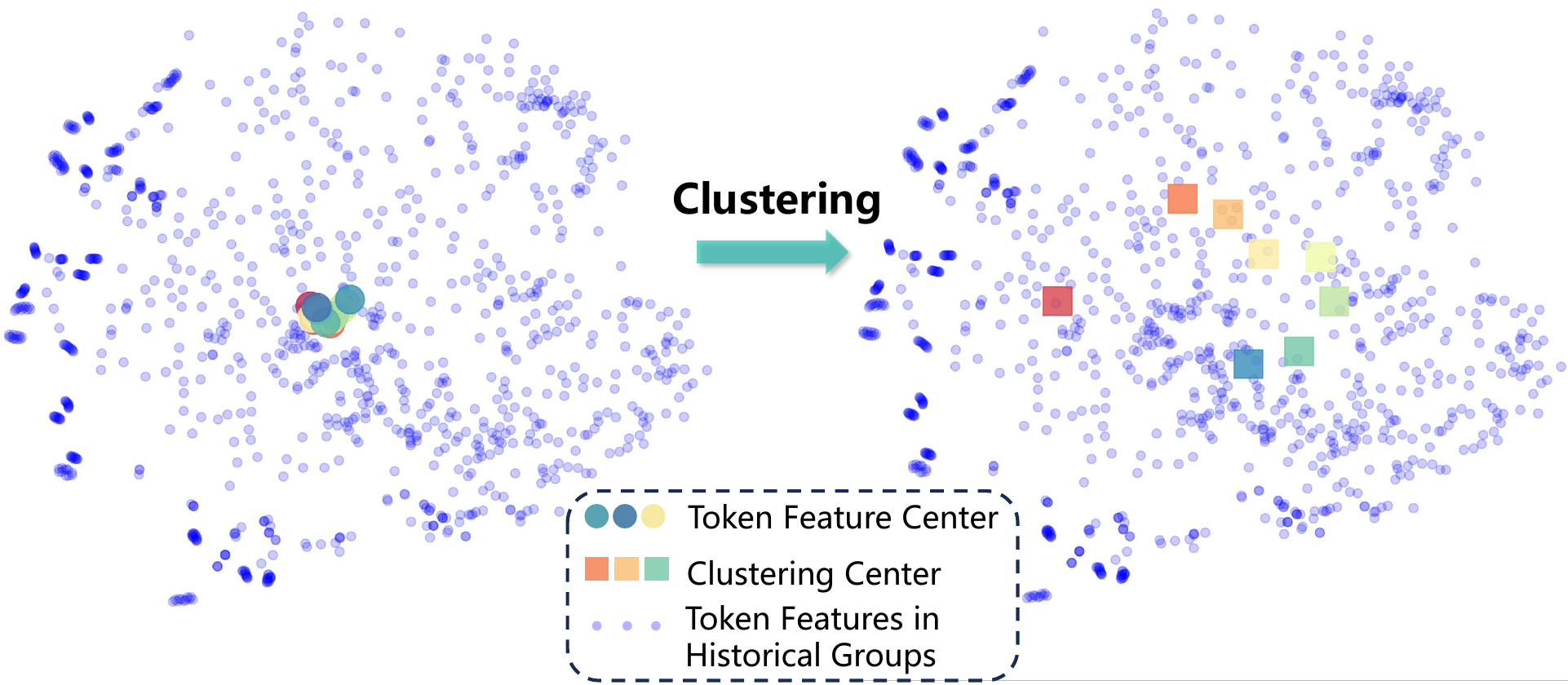}
\caption{\textbf{Visualization of the clustering within S-T Container.} The left is the original distribution of token features belonging to previous (historical) groups. The right visualizes the clustering centers after clustering. The clustering centers preserve the diversity of spatial-temporal information within token features.}\label{tsne}
\end{figure}

\textbf{Visualization of Clustering within S-T Container.} The visualization validates the feature update principle in our S-T container. We first plot the distribution of original token features from previous groups using PCA and T-SNE, marking their centers (left of Figure.~\ref{tsne}). After clustering these features via our S-T container, we visualize the cluster centers (right of Figure.~\ref{tsne}). Initially concentrated token feature centers indicate many token features sharing similarities, while the well-separated clustering centers after clustering preserve diverse spatial-temporal information within token features. By merging similar token features, S-T container updates token features in historical features into features with informative spatial-temporal representations. This effectively models long-term dependencies to guide predictions for subsequent groups.

\section{Applications}
\begin{figure}[t]%
\centering
\includegraphics[width=0.49\textwidth]{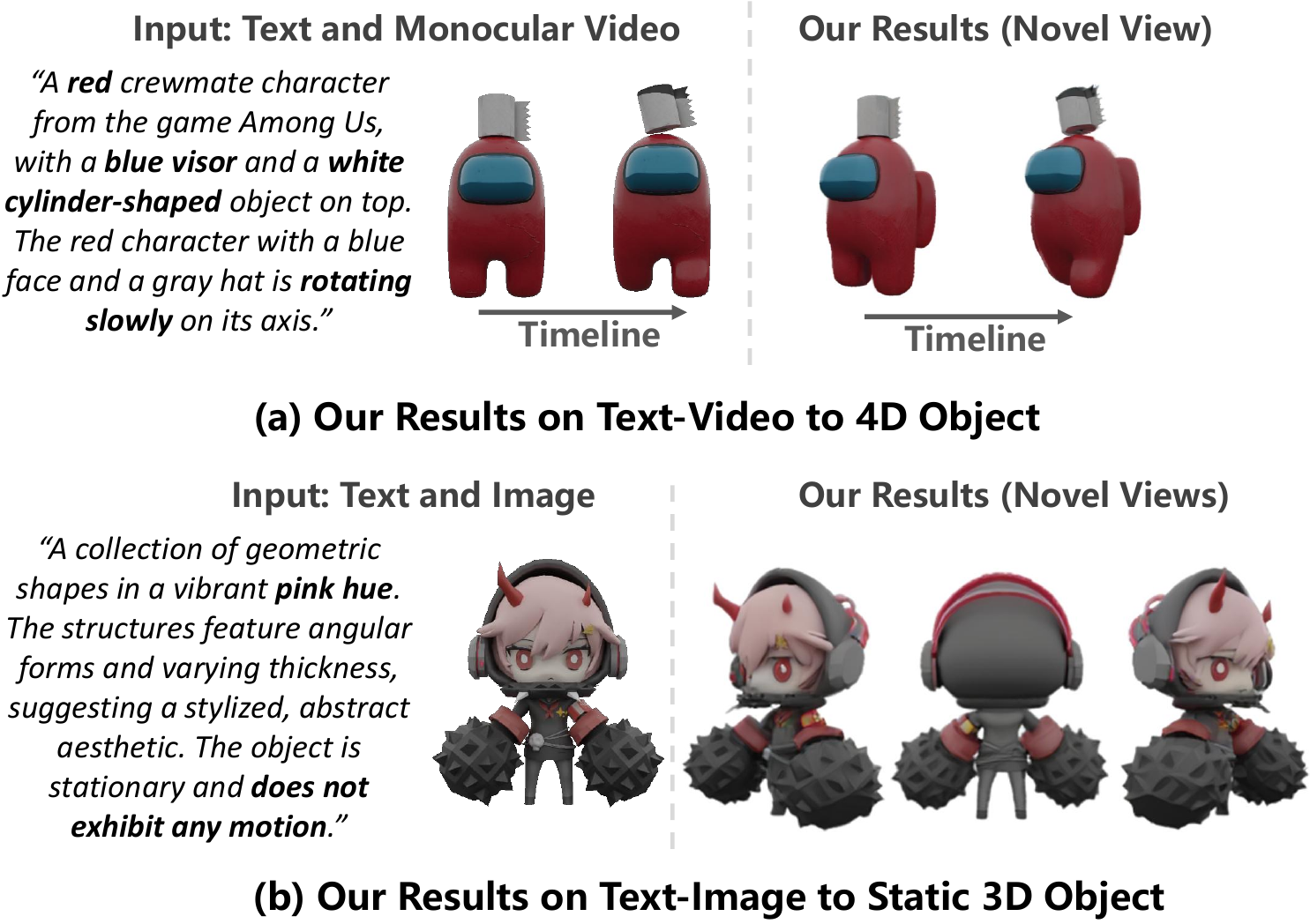}
\caption{\textbf{Visualization of (a) our text-video to 4D object generation and (b) our text-image to static 3D object generation.} Our approach is capable of generating a 4D object with temporal-spatial consistency and a 3D object with multi-view consistency.}\label{application}
\end{figure}

In Sec.~\ref{exps}, we verify our 4DSTAR is effective at generating 4D objects according to the input video. Additionally, our 4DSTAR supports the text prompt and corresponding video as input, as shown in Figure.~\ref{application} (a). Furthermore, our 4DSTAR can generate static 3D objects maintaining multi-view consistency when input text prompt and reference image, as shown in Figure.~\ref{application} (b).

\section{Conclusion}

In this paper, we propose a novel feed-forward Spatial-Temporal State Propagation Autoregressive Model named 4DSTAR, which generates temporal-spatial consistency 4D objects. Its STAR models long-term dependencies across generation results from previous timesteps via spatial-temporal state propagation, and leverages dependencies to guide generation at next timestep. A key component is the spatial-temporal container, integrated within STAR, which propagates prediction information from previous groups to enforce spatial-temporal consistency. Furthermore,  4D VQ-VAE is designed to decode the tokens predicted by STAR into temporally coherent dynamic 3D Gaussians. Overall, our method can generate high-quality 4D objects with temporal-spatial consistency.
{
    \small
    \bibliographystyle{ieeenat_fullname}
    \bibliography{ref}
}

\end{document}